\documentclass[journal]{IEEEtran}

\usepackage{amsmath}
\usepackage{amssymb}
\usepackage{pifont}
\usepackage{subfigure}
\usepackage{graphicx}
\usepackage{url}
\usepackage{cite}
\usepackage{multirow}
\usepackage{makecell}
\usepackage{booktabs}
\usepackage{bm}
\usepackage{graphics} 
\usepackage{epsfig}
\usepackage{balance}
\usepackage[ruled,linesnumbered]{algorithm2e}
\usepackage{longtable}
\usepackage{colortbl}
\usepackage{float}
\usepackage{color,xcolor}
\definecolor{light-gray}{gray}{0.9}
\definecolor{mycolor}{RGB}{0,0,0}
\usepackage{tikz}
\usepackage[colorlinks,linkcolor=blue,citecolor=blue,urlcolor=black]{hyperref}

\title{LoRA Patching: Exposing the Fragility of Proactive Defenses against Deepfakes}

\author{
Zuomin Qu\textsuperscript{1,2}, 
Yimao Guo\textsuperscript{1}, 
Qianyue Hu\textsuperscript{1}, 
Wei Lu\textsuperscript{1}* \\[0.5em]
\textsuperscript{1}School of Computer Science and Engineering, Sun Yat-sen University, Guangzhou 510006, China \\
\textsuperscript{2}Electric Power Research Institute, China Southern Power Grid Company Ltd., Guangzhou 510000, China \\
\texttt{quzuomin@csg.cn; guoym39, huqy56@mail2.sysu.edu.cn; luwei3@mail.sysu.edu.cn}
\thanks{*Corresponding author.}
}

\begin{document}

\maketitle

\begin{tikzpicture}[remember picture,overlay]
\node[anchor=south west,xshift=0.5cm,yshift=0.5cm] at (current page.south west) {%
    \footnotesize This work has been submitted to the IEEE for possible publication. Copyright may be transferred without notice, after which this version may no longer be accessible.
};
\end{tikzpicture}

\begin{abstract}
    Deepfakes pose significant societal risks, motivating the development of proactive defenses that embed adversarial perturbations in facial images to prevent manipulation. However, in this paper, we show that these preemptive defenses often lack robustness and reliability. We propose a novel approach, Low-Rank Adaptation (LoRA) patching, which injects a plug-and-play LoRA patch into Deepfake generators to bypass state-of-the-art defenses. A learnable gating mechanism adaptively controls the effect of the LoRA patch and prevents gradient explosions during fine-tuning. We also introduce a Multi-Modal Feature Alignment (MMFA) loss, encouraging the features of adversarial outputs to align with those of the desired outputs at the semantic level. Beyond bypassing, we present defensive LoRA patching, embedding visible warnings in the outputs as a complementary solution to mitigate this newly identified security vulnerability. With only 1,000 facial examples and a single epoch of fine-tuning, LoRA patching successfully defeats multiple proactive defenses. These results reveal a critical weakness in current paradigms and underscore the need for more robust Deepfake defense strategies. Our code is available at \url{https://github.com/ZOMIN28/LoRA-Patching}.
\end{abstract}


\vspace{-10pt}
\section{Introduction}
\begin{figure}[ht]
      \centering
      \includegraphics[width=1.0\linewidth]{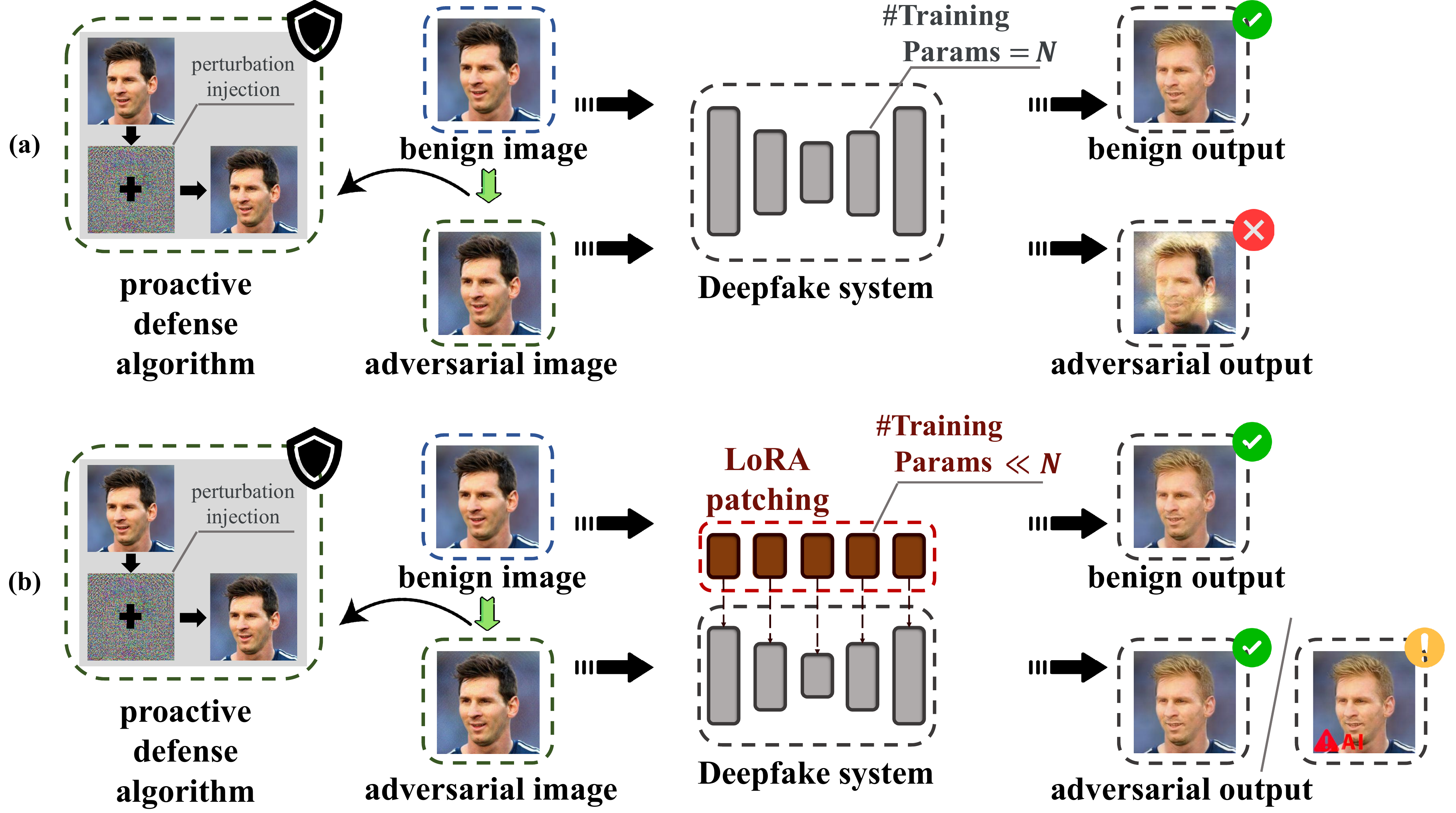}
      \caption{Illustration of proactive Deepfake defenses and LoRA patching bypass. (a) Proactive defenses embed invisible adversarial perturbations into images to disrupt Deepfakes; (b) LoRA patching inserts LoRA blocks into each linear, convolutional, and transposed convolutional layer, introducing only a lightweight set of additional parameters relative to the full model, while preventing proactive defenses from disrupting manipulated images and preserving manipulation of benign ones.}
      \label{fig:intro}
\end{figure}

\IEEEPARstart{R}{ecent} advances in AI generative technology have enabled the creation of highly convincing Deepfake facial images~\cite{li2023agiqa,li2025big, zhu2024stableswap,wang2024efficient}. The malicious use of this technology poses serious risks to individual rights and social stability~\cite{chu2022protecting,wang2024dual,huang2023dodging,liao2023famm}. To address it, several studies have proposed proactive defenses against Deepfakes~\cite{ruiz2020disrupting,huang2021initiative,huang2022cmua,zhai2023defending,qu2024df,xu2024robust,qu2025id,zhang2025synth,mi2025preemptive}, exploiting the vulnerability of AI models to adversarial attacks\cite{ruiz2020disrupting,madry2017towards,guo2025anti}. As illustrated in Fig.~\hyperref[fig:intro]{\ref*{fig:intro}(a)}, these defenses embed adversarial perturbations into images in advance. When such adversarial images are fed into Deepfake generators, the resulting outputs display visually unnatural distortions, thereby reducing their credibility; for example, editing the hair color of Lionel Messi's photo leads to severe artifacts under proactive defenses.

\begin{figure*}[ht]
      \centering
      \includegraphics[width=0.95\linewidth]{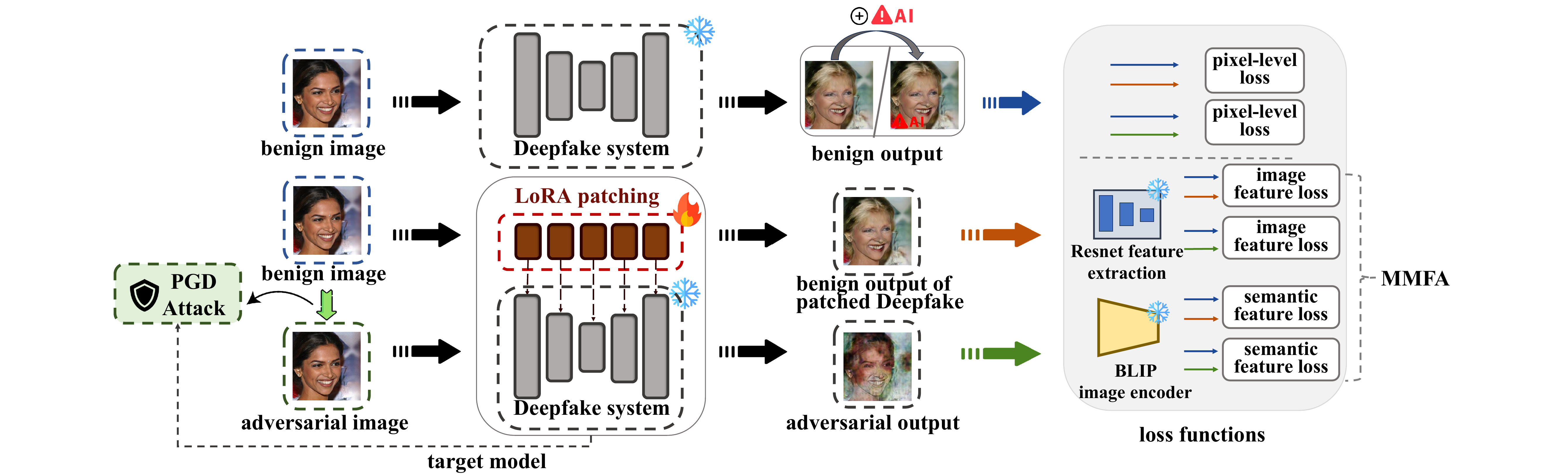}
      \caption{Illustration of the LoRA patching fine-tuning process. A bi-level min-max optimization approach based on adversarial training is proposed for fine-tuning, where the inner maximization uses PGD to generate adversarial examples with the \textbf{current} patched deepfake as the target model.}
      \label{fig:pipeline}
\end{figure*}

\begin{figure}[ht]
      \centering
      \includegraphics[width=1.0\linewidth]{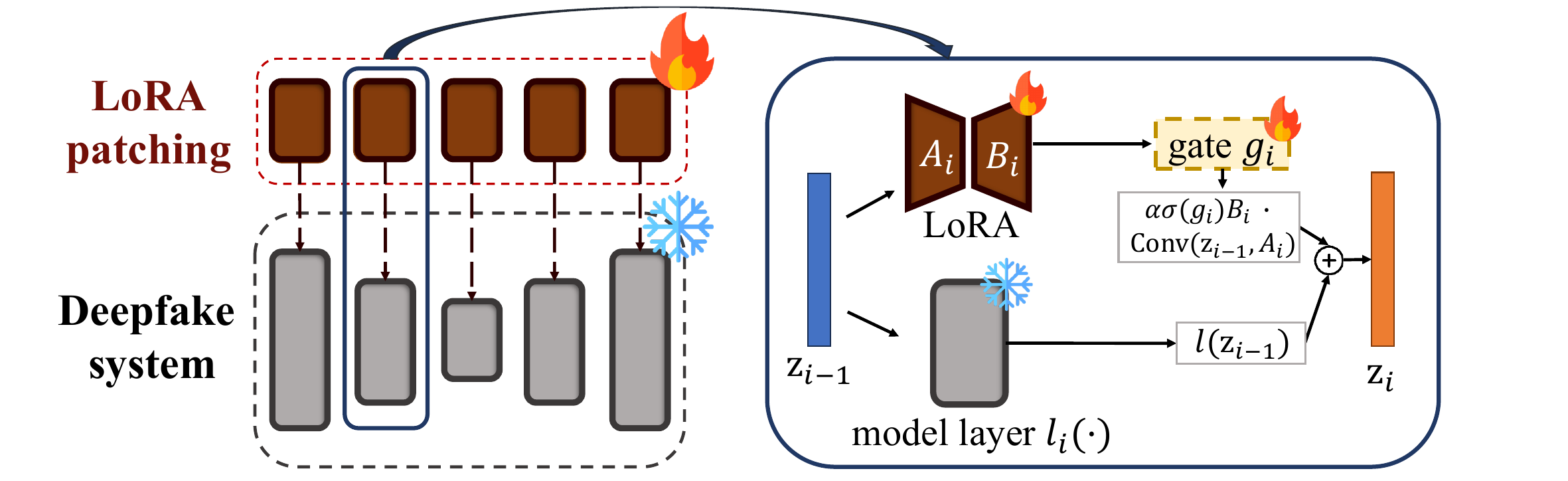}
      \caption{Illustration of LoRA patch embedding. A pair of LoRA blocks is inserted into each convolutional and deconvolutional layer of the Deepfake model to adjust the output. Each layer further includes a learnable gating parameter that adaptively trades off the patch’s influence.}
      \label{fig:gate}
\end{figure}

However, we identify a security vulnerability in these proactive defenses, as they can be bypassed using carefully designed patches. Specifically, we propose Low-Rank Adaptation (LoRA)~\cite{hu2022lora} patching for well-trained Deepfakes, which neutralize adversarial perturbations by injecting LoRA blocks into each layer in Deepfake generators. As illustrated in Fig.~\hyperref[fig:intro]{\ref*{fig:intro}(b)}, this approach preserves the generator's performance on adversarial images while maintaining normal functionality on benign examples. To stabilize fine-tuning and align features of benign and adversarial outputs, we introduce a learnable gating mechanism and a Multi-Modal Feature Alignment (MMFA) loss. Beyond bypassing proactive defenses, we further propose defensive LoRA patching, embedding visible warning watermarks in generated outputs as a complementary measure to mitigate this security vulnerability. 

Compared with existing adversarial defense methods, including image preprocessing~\cite{dziugaite2016study,jia2019comdefend,zhang2021defense} and conventional adversarial training~\cite{ruiz2020disrupting,luo2024magat,huang2023fast}, the proposed LoRA patching introduces a novel adversarial training-based patching paradigm that achieves stronger robustness, plug-and-play integration, and significantly lower computational cost. A related work, CAT~\cite{peng2025cat}, investigates the robustness of diffusion-based generative models against proactive perturbations via LoRA-based adversarial training. In contrast, our work designs LoRA patches specifically for convolutional architectures in GAN-based Deepfake systems and further explores their potential to actively embed visible signals in the generated outputs.

Our main contributions are summarized as follows:
\begin{enumerate}
    \item We uncover the vulnerabilities of state-of-the-art proactive Deepfake defenses and further propose LoRA patching to bypass these defenses.
    \item We introduce a learnable gating mechanism to stabilize LoRA fine-tuning and an MMFA loss to efficiently restore disrupted outputs.
    \item Extensive experiments demonstrate that LoRA patching effectively bypasses existing proactive defenses while preserving high-quality face forgeries.
\end{enumerate}

\vspace{-10pt}
\section{Method}  \label{Method}

\subsection{Threat Model}

\subsubsection{System Overview}
Given an input face $x$, a Deepfake model $\mathcal{M}: X \to Y$ produces a manipulated output $y=\mathcal{M}(x)$ in real time. In practice, attackers can easily access open-source face-swapping and attribute-editing models to perform forgery with minimal setup \cite{qu2025id}.

\subsubsection{Defender Model}
We consider proactive defenses with white-box access to $\mathcal{M}$. The defender embeds a bounded perturbation $\delta$ into the input, generating $\hat{x}=x+\delta$, where $\delta$ is optimized to maximize the divergence between $\mathcal{M}(x)$ and $\mathcal{M}(\hat{x})$, thereby disrupting the forged output.

\subsubsection{Attacker Model}
The attacker seeks to neutralize the defender’s perturbation while preserving high-quality forgery. We assume the attacker can load the public model $\mathcal{M}$ and insert lightweight LoRA modules without altering original weights, enabling a parameter-efficient adaptation trained on both $x$ and $\hat{x}$ at very low cost. The patched model $\mathcal{M}_p$ is expected to map both inputs to the desired fake domain $Y$.

\vspace{-10pt}
\subsection{LoRA Patching}
\subsubsection{Overview}
The LoRA patching fine-tunes a small set of trainable matrices inserted into the pre-trained model $\mathcal{M}_o$. The objective is to minimize the discrepancy between the outputs of the patched model $\mathcal{M}_p$ and the original output $y=\mathcal{M}_o(x)$ for both benign inputs $x$ and adversarial inputs $\hat{x}$ :
\begin{equation}
\min_{\theta_{p}} \big( \mathcal{L}(\mathcal{M}_{p}(x), y) + \mathcal{L}(\mathcal{M}_{p}(\hat{x}), y) \big),
\label{eq2}
\end{equation}
where $\theta_{p}$ denotes the LoRA patch parameters, and $\mathcal{L}$ is the fine-tuning loss function (introduced in Section~\ref{Loss Function}).

A critical aspect of this process is the generation of adversarial examples $\hat{x}$, which can be divided into two scenarios: 1) \textbf{the standard scenario}, where the defender generates adversarial perturbations using the original Deepfake $\mathcal{M}_{o}$ as the target model; and 2) \textbf{the leakage scenario}, where the patched Deepfake $\mathcal{M}_{p}$ is exposed, allowing the defender to access its parameters and generate perturbations for the entire model. To ensure the robustness of LoRA patching under both scenarios, we adopt a bi-level min-max optimization paradigm based on adversarial training. The inner maximization employs PGD~\cite{madry2017towards} to generate adversarial examples with the \textit{current} patched model as the target, while the outer minimization updates the LoRA patch parameters to minimize the loss:
\begin{equation}
\min_{\theta_{p}}\,
\!\Big[
    \underbrace{\mathcal{L}(\mathcal{M}_{p}(x),y)}_{\text{benign consistency}}
    +\underbrace{\max_{\|\delta\|_\infty\le\epsilon}\mathcal{L}(\mathcal{M}_{p}(x+\delta),y)}_{\text{adversarial consistency}}
\Big],
\label{eq3}
\end{equation}
where $y = \mathcal{M}_{o}(x)$, and $\mathcal{M}_{p}$ is the current patched Deepfake.

\subsubsection{Learnable Gating Mechanism} \label{Learnable Gating Mechanism}
For each convolutional and transposed convolutional layer in the original Deepfake model, we insert a pair of LoRA blocks to modulate its output. Fig.~\ref{fig:gate} illustrates the embedding of a LoRA patch within a given layer. Specifically, for a layer $l_i$ in the Deepfake network, the associated LoRA matrices $A_i$ and $B_i$ generate a patch output $B_i A_i z_{i-1}$, where $z_{i-1}$ denotes the input to this layer.To stabilize fine-tuning, we introduce a learnable gating mechanism at each layer, parameterized by $g_i$, which is passed through a Sigmoid activation $\sigma(\cdot)$ and jointly updated with the LoRA matrices. This design allows the network to adaptively control the contribution of the LoRA patch at the current layer. Consequently, the output of a patched layer is formulated as
$z_i = l_i(z_{i-1}) + \alpha \sigma(g_i) B_i \cdot \text{Conv}(z_{i-1},A_i),$
where $\alpha$ is a prior hyperparameter that balances the influence of the LoRA patch.
For linear layers, the LoRA modules follow the standard low-rank decomposition and scaling strategy proposed in \cite{hu2022lora}.
The parameters of the LoRA patch, i.e., $\theta_p$ in Eq.~\ref{eq2}, thus consist of the set $\{g, A, B\}$.

\subsubsection{Defensive LoRA Patching}
Motivated by the identified security risks, we further explore the use of LoRA patching for defensive purposes. The patched model is trained to embed a visible warning mark in its outputs explicitly, indicating potential manipulation. As shown in Fig.~\ref{fig:pipeline}, we generate a target $y_w=\mathcal{M}_o(x)+w$ by adding a watermark $w$ to the benign output. During fine-tuning, we replace $y$ with $y_w$ in Eq.~\ref{eq2}, enabling the patched model to map both $x$ and $\hat{x}$ to the watermarked image domain.

\vspace{-10pt}
\subsection{Loss Function} \label{Loss Function}

\begin{figure}[t]
      \centering
      \includegraphics[width=1.0\linewidth]{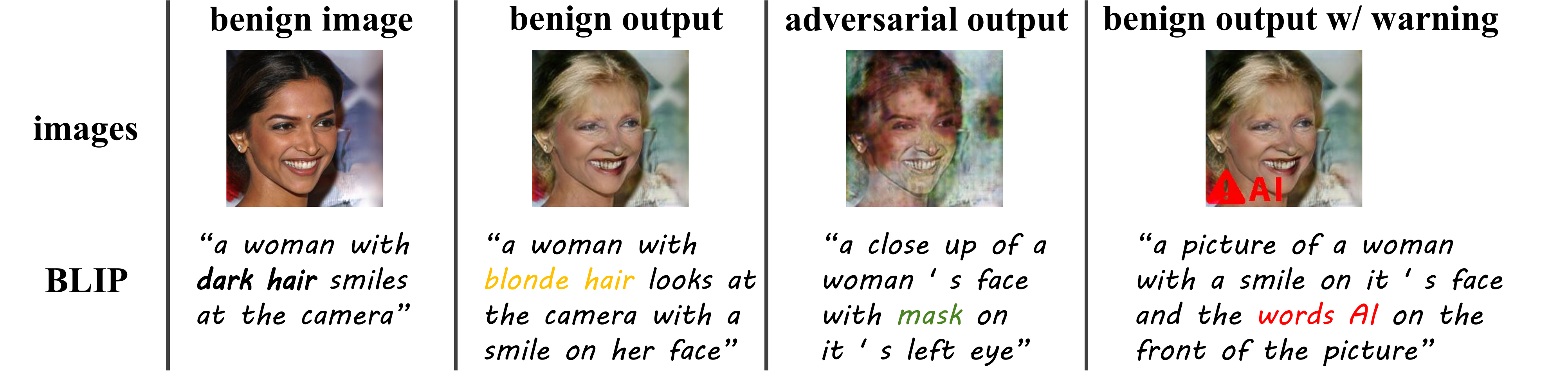}
      \caption{Illustration of textual descriptions generated by BLIP's vision–language encoder~\cite{li2022blip} for the images.}
      \label{fig:blip}
\end{figure}

\begin{table}[t]
    \centering
    \caption{Quantitative results of bypassing proactive defenses in general scenarios.}
    \label{tab:mainresult}
    \resizebox{1.0\columnwidth}{!}{
    \begin{tabular}{c|c|ccccccccc}
    \toprule
    \multirow{2}{*}{Defenses} 
    &\multirow{2}{*}{Methods} 
    &\multicolumn{3}{c}{StarGAN\cite{choi2018stargan}}  &\multicolumn{3}{c}{AttGAN\cite{he2019attgan}} 
    &\multicolumn{3}{c} {HiSD\cite{li2021image}} \\
  \cmidrule(l){3-5}  \cmidrule(l){6-8}  \cmidrule(l){9-11}& 
  &$L_{2}$$\downarrow$ &SSIM$\uparrow$  &DSR$\downarrow$ 
  &$L_{2}$$\downarrow$ &SSIM$\uparrow$  &DSR$\downarrow$ 
  &$L_{2}$$\downarrow$ &SSIM$\uparrow$  &DSR$\downarrow$ \\
    \midrule
    
    \multirow{6}{*}{Disrupting\cite{ruiz2020disrupting}} 
    &No bypassing           
    &1.369 &0.230 &1.000 &0.243 &0.667 &0.952 &0.428 &0.478 &1.000\\
    \cmidrule(lr){2-11}
    &JC~\cite{dziugaite2016study}          
    &0.039 &0.909 &0.276 &0.073 &0.846 &0.444 &0.011 &\textbf{0.943} &0.020\\
    &ComDefend~\cite{jia2019comdefend}          
    &0.037 &0.837 &0.219 &0.024 &0.857 &0.076 &0.034 &0.859 &0.217\\
    &RI~\cite{zhang2021defense}          
    &0.031 &0.887 &0.124 &0.030 &0.865 &0.104 &0.021 &0.897 &0.067\\
    &\cellcolor{light-gray}Ours     
    &\cellcolor{light-gray}\textbf{0.018} &\cellcolor{light-gray}\textbf{0.911} &\cellcolor{light-gray}\textbf{0.026} &\cellcolor{light-gray}\textbf{0.009} &\cellcolor{light-gray}\textbf{0.933} &\cellcolor{light-gray}\textbf{0.008} &\cellcolor{light-gray}\textbf{0.007} &\cellcolor{light-gray}0.942 &\cellcolor{light-gray}\textbf{0.011}\\
    \cmidrule(lr){1-11}   
    
    \multirow{6}{*}{PG\cite{huang2021initiative}} 
    &No bypassing             
    &2.002 &0.153 &1.000 &0.132 &0.771 &0.454 &0.425 &0.515 &1.000\\
    \cmidrule(lr){2-11}
    &JC~\cite{dziugaite2016study}           
    &0.044 &0.855 &0.312 &0.052 &0.870 &0.236 &0.014 &\textbf{0.965} &\textbf{0.017}\\
    &ComDefend~\cite{jia2019comdefend}          
    &0.037 &0.838 &0.224 &0.024 &0.854 &0.073 &0.035 &0.856 &0.221\\
    &RI~\cite{zhang2021defense}          
    &0.031 &0.898 &0.142 &0.031 &0.867 &0.134 &0.023 &0.893 &0.103\\
    &\cellcolor{light-gray}Ours     
    &\cellcolor{light-gray}\textbf{0.017} &\cellcolor{light-gray}\textbf{0.917} &\cellcolor{light-gray}\textbf{0.024} &\cellcolor{light-gray}\textbf{0.007} &\cellcolor{light-gray}\textbf{0.938} &\cellcolor{light-gray}\textbf{0.006} &\cellcolor{light-gray}\textbf{0.012} &\cellcolor{light-gray}0.937 &\cellcolor{light-gray}0.020\\
    \cmidrule(lr){1-11}
    
    \multirow{6}{*}{CMUA\cite{huang2022cmua}} 
    &No bypassing         
    &0.173 &0.625 &1.000 &0.038 &0.779 &0.172 &0.070 &0.782 &0.571\\
    \cmidrule(lr){2-11}
    &JC~\cite{dziugaite2016study}           
    &\textbf{0.005} &\textbf{0.951} &\textbf{0.002} &0.009 &0.934 &0.012 &\textbf{0.003} &0.945 &\textbf{0.008}\\
    &ComDefend~\cite{jia2019comdefend}          
    &0.037 &0.837 &0.219 &0.022 &0.860 &0.052 &0.033 &0.859 &0.220\\
    &RI~\cite{zhang2021defense}         
    &0.033 &0.893 &0.162 &0.022 &0.889 &0.032 &0.020 &0.897 &0.063\\
    &\cellcolor{light-gray}Ours     
    &\cellcolor{light-gray}0.017 &\cellcolor{light-gray}0.916 &\cellcolor{light-gray}0.024 &\cellcolor{light-gray}\textbf{0.008} &\cellcolor{light-gray}\textbf{0.936} &\cellcolor{light-gray}\textbf{0.004} &\cellcolor{light-gray}0.009 &\cellcolor{light-gray}\textbf{0.952} &\cellcolor{light-gray}\textbf{0.008}\\
    \cmidrule(lr){1-11}
    
    \multirow{6}{*}{DF-RAP\cite{qu2024df}} 
    &No bypassing
    &0.368 &0.517 &1.000 &0.237 &0.702 &0.914 &0.207 &0.623 &0.997\\
    \cmidrule(lr){2-11}
    &JC~\cite{dziugaite2016study}           
    &0.154 &0.766 &0.952 &0.157 &0.773 &0.736 &0.025 &0.936 &0.127\\
    &ComDefend~\cite{jia2019comdefend}          
    &0.037 &0.836 &0.224 &0.025 &0.854 &0.086 &0.039 &0.858 &0.252\\
    &RI~\cite{zhang2021defense}          
    &0.031 &0.895 &0.126 &0.037 &0.863 &0.192 &0.028 &0.880 &0.146\\
    &\cellcolor{light-gray}Ours     
    &\cellcolor{light-gray}\textbf{0.018} &\cellcolor{light-gray}\textbf{0.910} &\cellcolor{light-gray}\textbf{0.026} &\cellcolor{light-gray}\textbf{0.008} &\cellcolor{light-gray}\textbf{0.933} &\cellcolor{light-gray}\textbf{0.002} &\cellcolor{light-gray}\textbf{0.011} &\cellcolor{light-gray}\textbf{0.939} &\cellcolor{light-gray}\textbf{0.033}\\
    \cmidrule(lr){1-11}

    \multirow{6}{*}{RAW\cite{xu2024robust}} 
    &No bypassing
    &1.287 &0.156 &1.000 &0.218 &0.590 &0.939 &0.412 &0.392 &1.000\\
    \cmidrule(lr){2-11}
    &JC~\cite{dziugaite2016study}           
    &0.073 &0.798 &0.632 &0.079 &0.767 &0.478 &0.013 &\textbf{0.939} &0.028\\
    &ComDefend~\cite{jia2019comdefend}          
    &0.037 &0.745 &0.227 &0.024 &0.757 &0.066 &0.033 &0.770 &0.205\\
    &RI~\cite{zhang2021defense}          
    &0.029 &0.845 &0.112 &0.029 &0.806 &0.108 &0.021 &0.844 &0.048\\
    &\cellcolor{light-gray}Ours     
    &\cellcolor{light-gray}\textbf{0.017} &\cellcolor{light-gray}\textbf{0.868} &\cellcolor{light-gray}\textbf{0.024} &\cellcolor{light-gray}\textbf{0.008} &\cellcolor{light-gray}\textbf{0.889} &\cellcolor{light-gray}\textbf{0.005} &\cellcolor{light-gray}\textbf{0.009} &\cellcolor{light-gray}0.912 &\cellcolor{light-gray}\textbf{0.017}\\
    \cmidrule(lr){1-11}

    \multirow{6}{*}{ID-Guard\cite{qu2025id}} 
    &No bypassing
    &0.388 &0.301 &1.000 &0.039 &0.445 &0.122 &0.058 &0.539 &0.562\\
    \cmidrule(lr){2-11}
    &JC~\cite{dziugaite2016study}           
    &0.043 &0.860 &0.307 &0.013 &0.716 &0.012 &0.019 &0.930 &0.022\\
    &ComDefend~\cite{jia2019comdefend}          
    &0.038 &0.738 &0.233 &0.023 &0.758 &0.049 &0.034 &0.766 &0.212\\
    &RI~\cite{zhang2021defense}          
    &0.030 &0.834 &0.126 &0.020 &0.831 &0.032 &0.019 &0.839 &0.045\\
    &\cellcolor{light-gray}Ours     
    &\cellcolor{light-gray}\textbf{0.017} &\cellcolor{light-gray}\textbf{0.863} &\cellcolor{light-gray}\textbf{0.026} &\cellcolor{light-gray}\textbf{0.007} &\cellcolor{light-gray}\textbf{0.892} &\cellcolor{light-gray}\textbf{0.002} &\cellcolor{light-gray}\textbf{0.009} &\cellcolor{light-gray}\textbf{0.909} &\cellcolor{light-gray}\textbf{0.012}\\
    \bottomrule
    \end{tabular}}
\end{table}

Our goal is to update the embedded LoRA patches so that the patched Deepfake model $\mathcal{M}_{p}$ maps both benign $x$ and adversarial $\hat{x}$ images to the target output $y$. We first adopt a pixel-level loss for per-pixel consistency:
\begin{equation}
\mathcal{L}_{\text{pix}} = \Vert \mathcal{M}_{p}(x) - y \Vert_{2}^{2} + \Vert \mathcal{M}_{p}(\hat{x}) - y \Vert_{2}^{2}.
\end{equation}

Next, to capture fine-grained visual discrepancies beyond pixels, we design a Multi-Modal Feature Alignment (MMFA) loss. Deepfakes often involve subtle attribute manipulations (e.g., modifying facial regions such as the mouth or eyes), which pixel-level losses fail to capture. To better restore these details, we adopt an image feature loss:
\begin{equation}
\mathcal{L}_{\text{img}} = \Vert \mathcal{F}(\mathcal{M}_{p}(x)) - \mathcal{F}(y) \Vert_{2}^{2} + \Vert \mathcal{F}(\mathcal{M}_{p}(\hat{x})) - \mathcal{F}(y) \Vert_{2}^{2},
\end{equation}
where $\mathcal{F}$ is a pretrained ResNet-50~\cite{he2016deep} truncated before the final classification layer. As illustrated in Fig.~\ref{fig:blip}, we leverage BLIP’s vision-language encoder~\cite{li2022blip} to generate textual descriptions of the images. These descriptions show significant semantic discrepancies between pre- and post-editing images, as well as between protected and unprotected outputs. To address this gap, we introduce a semantic feature loss to align semantic representations:
\begin{equation}
\mathcal{L}_{\text{sem}} = \Vert E(\mathcal{M}_{p}(x)) - E(y) \Vert_{2}^{2} + \Vert E(\mathcal{M}_{p}(\hat{x})) - E(y) \Vert_{2}^{2},
\end{equation}
where $E$ denotes the pretrained BLIP image encoder, which effectively extracts semantic-related image features. This loss is critical for defensive LoRA patching, since the visible warning watermark exhibits prominent semantic characteristics.

Finally, the overall training objective is formulated as:
\begin{equation}
\mathcal{L} = \mathcal{L}_{\text{pix}} + \lambda_{1} \mathcal{L}_{\text{img}} + \lambda_{2} \mathcal{L}_{\text{sem}},
\end{equation}
where $\lambda_{1}$ and $\lambda_{2}$ balance each item.

\begin{table}[t]
    \centering
    \caption{Quantitative results on the impact of bypass mechanisms on benign outputs.}
    \label{tab:ben_vis}
    \resizebox{1.0\columnwidth}{!}{
    \begin{tabular}{c|ccccccccc}
    \toprule
    \multirow{2}{*}{Methods} 
    &\multicolumn{3}{c}{StarGAN\cite{choi2018stargan}}  &\multicolumn{3}{c}{AttGAN\cite{he2019attgan}} 
    &\multicolumn{3}{c} {HiSD\cite{li2021image}} \\
      \cmidrule(l){2-4}  \cmidrule(l){5-7}  \cmidrule(l){8-10}& 
      FID$\downarrow$ &SSIM$\uparrow$ &BRISQUE$\downarrow$ 
      &FID$\downarrow$ &SSIM$\uparrow$ &BRISQUE$\downarrow$  
      &FID$\downarrow$ &SSIM$\uparrow$ &BRISQUE$\downarrow$ \\
    \midrule
    JC~\cite{dziugaite2016study}          
    &8.004 &0.991 &18.503 &7.837 &0.998 &16.243 &7.877 &0.992 &12.208\\
    ComDefend~\cite{jia2019comdefend}          
    &12.601 &0.838 &49.550 &14.617 &0.857 &55.723 &14.819 &0.860 &45.257\\
    RI~\cite{zhang2021defense}          
    &10.521 &0.892 &35.016 &12.840 &0.886 &35.079 &12.689 &0.906 &16.691\\
    \cellcolor{light-gray}Ours     
    &\cellcolor{light-gray}9.675 &\cellcolor{light-gray}0.912 &\cellcolor{light-gray}24.460 &\cellcolor{light-gray}9.269 &\cellcolor{light-gray}0.936 &\cellcolor{light-gray}25.129 &\cellcolor{light-gray}8.887 &\cellcolor{light-gray}0.951 &\cellcolor{light-gray}15.795\\
    \bottomrule
    \end{tabular}}
\end{table}

\begin{table}[t]
    \centering
    \caption{Quantitative comparison of LoRA patching and adversarial training methods in bypassing Disrupting\cite{ruiz2020disrupting}.}
    \label{tab:with_AT}
    \resizebox{1.0\columnwidth}{!}{
    \begin{tabular}{c|ccc|cc}
        \toprule
        Methods &$L_{2}$$\downarrow$ &SSIM$\uparrow$ &DSR$\downarrow$ &Training Time (h) $\downarrow$ &\#Training Params (M) $\downarrow$\\
        \midrule
        GAT\cite{ruiz2020disrupting} &0.077 &0.818 &0.422 &5.245 &8.431\\
        MaGAT\cite{luo2024magat} &0.031 &0.892 &0.093  &5.370 &8.431\\
        \cellcolor{light-gray}Ours &\cellcolor{light-gray}\textbf{0.017} &\cellcolor{light-gray}\textbf{0.917} &\cellcolor{light-gray}\textbf{0.026} &\cellcolor{light-gray}\textbf{0.251} &\cellcolor{light-gray}\textbf{0.360}\\
        \bottomrule
    \end{tabular}}
\end{table}

\begin{figure}[t]
      \centering
      \setcounter{subfigure}{0}
      \subfigure [$L_{2}$.] {\includegraphics[width=.15\textwidth]{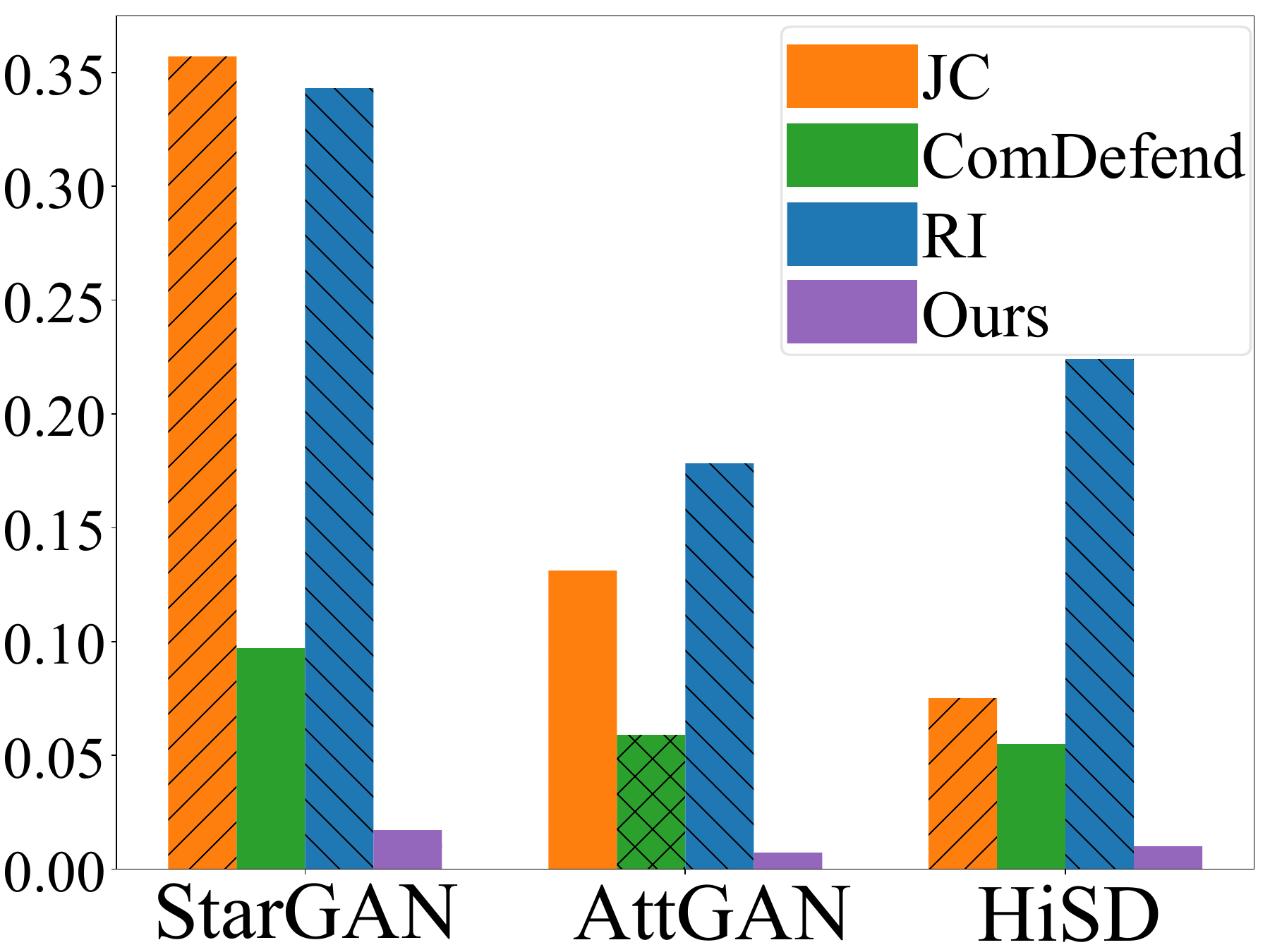}}
      \subfigure [SSIM.] {\includegraphics[width=.15\textwidth]{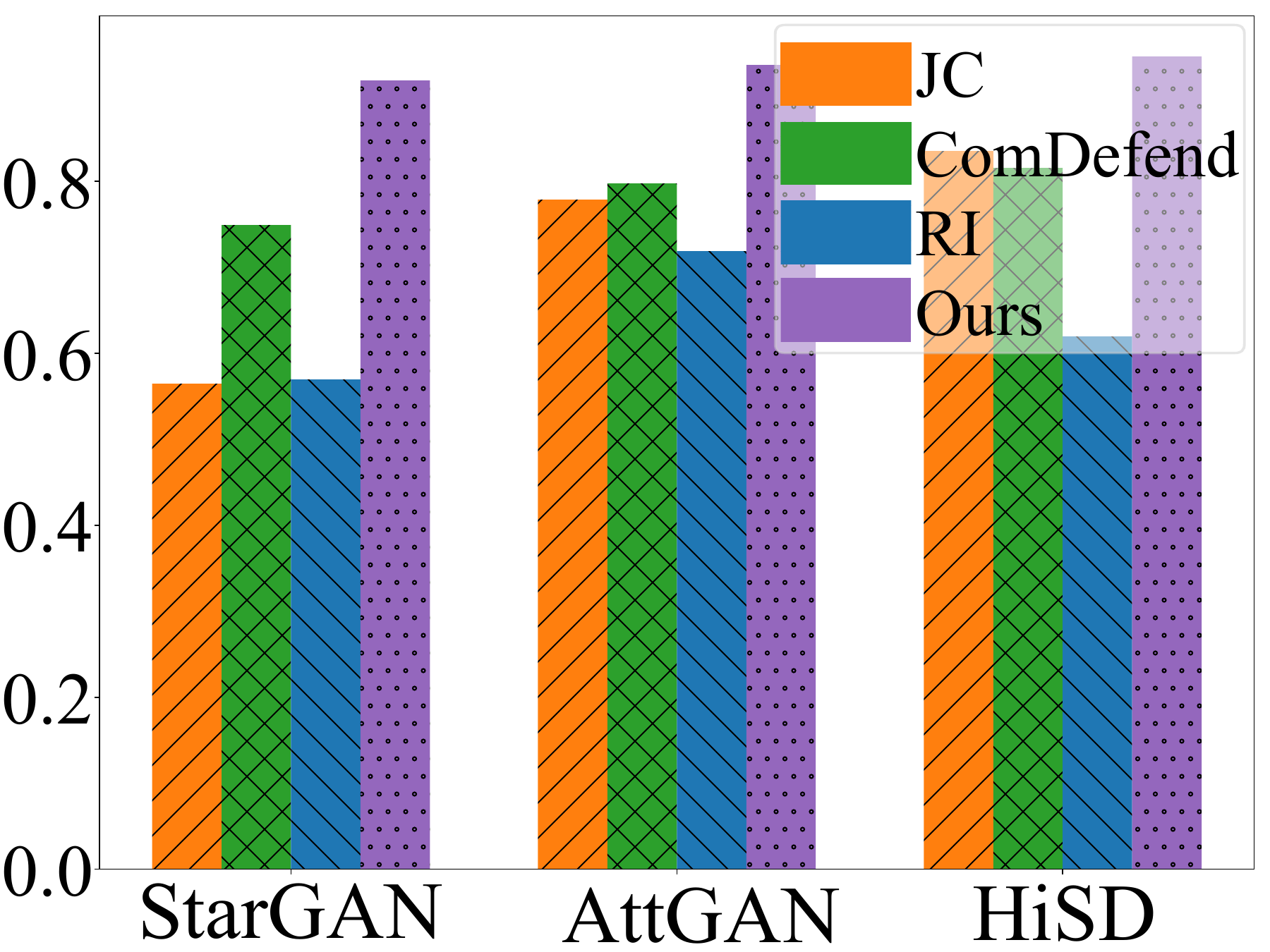}}
      \subfigure [DSR.] {\includegraphics[width=.15\textwidth]{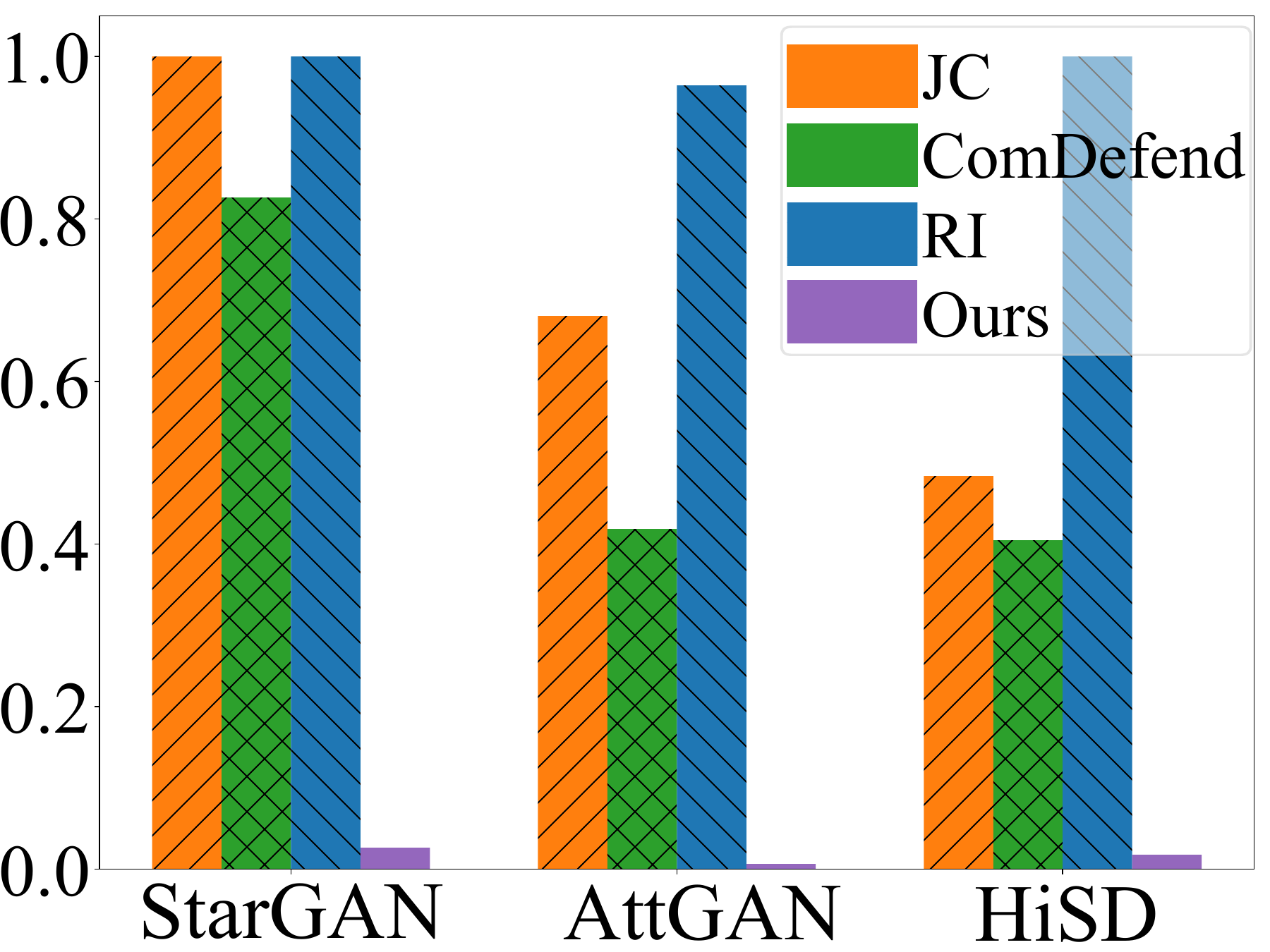}}
      \caption{Quantitative results of different methods bypassing proactive defenses in leakage scenarios.}
      \label{fig:leakage}
\end{figure}

\vspace{-10pt}

\section{Experiments} \label{Experiments}

\subsection{Experimental Setup}

\subsubsection{Datasets}
We conduct experiments on the CelebA~\cite{liu2015deep} dataset, which contains 202,599 facial images with attribute annotations. The first 1,000 images are used to fine-tune the LoRA patch.

\subsubsection{Baselines}
We select StarGAN\cite{choi2018stargan}, AttGAN~\cite{he2019attgan}, and HiSD~\cite{li2021image} as target Deepfake models, and employ four proactive defenses, including Disrupting~\cite{ruiz2020disrupting}, PG~\cite{huang2021initiative}, CMUA~\cite{huang2022cmua}, DF-RAP~\cite{qu2024df}, RAW~\cite{xu2024robust}, and ID-Guard~\cite{qu2025id} to generate adversarial perturbations. We compare the proposed LoRA patching with five representative adversarial defense methods, distinguishing preprocessing-based defenses (JC~\cite{dziugaite2016study}, ComDefend~\cite{jia2019comdefend}, RI~\cite{zhang2021defense}) from adversarial training-based defenses (GAT~\cite{ruiz2020disrupting}, MaGAT~\cite {luo2024magat}).

\subsubsection{Evaluation Metrics}
Following prior studies on proactive defenses~\cite{ruiz2020disrupting,qu2024df}, we measure L2 distance, SSIM~\cite{wang2004image}, and Defense Success Rate (DSR) to quantify defense bypassing, where lower defense metrics indicate better bypass performance. We include FID~\cite{heusel2017gans} and BRISQUE~\cite{mittal2012no} to assess the visual quality.

\subsubsection{Implementation details}
Images are resized to $256 \times 256$ and normalized to $[-1,1]$. Fine-tuning uses batch size = 4, and MMFA loss weights $\lambda_1 = \lambda_2 = 0.1$.

\vspace{-10pt}

\subsection{Comparison Results}
\subsubsection{Standard Scenario}
Under the standard scenario, the quantitative results for bypassing proactive defenses are reported in Table~\ref{tab:mainresult}. Compared with baselines, the proposed LoRA patching achieves the best performance, reducing the average DSR of the evaluated defenses from 83.8\% to 1.6\%. Owing to the vulnerability of adversarial perturbations to image compression, JC~\cite{dziugaite2016study} also performs well against general defenses but remains ineffective against the compression-robust DF-RAP~\cite{qu2024df} and RAW~\cite{xu2024robust}. Table~\ref{tab:ben_vis} further reports the quantitative impact on benign samples, showing that JC introduces the least distortion, while our method is second with only a marginal gap, yet delivers substantially stronger bypassing performance. This superior performance stems from the LoRA patching’s layer-wise adaptation and learnable gating mechanism, which allow it to selectively counteract adversarial perturbations without disrupting the generator’s normal mapping.

\subsubsection{Leakage Scenario}
\begin{figure}[t]
      \centering
      \includegraphics[width=1.\linewidth]{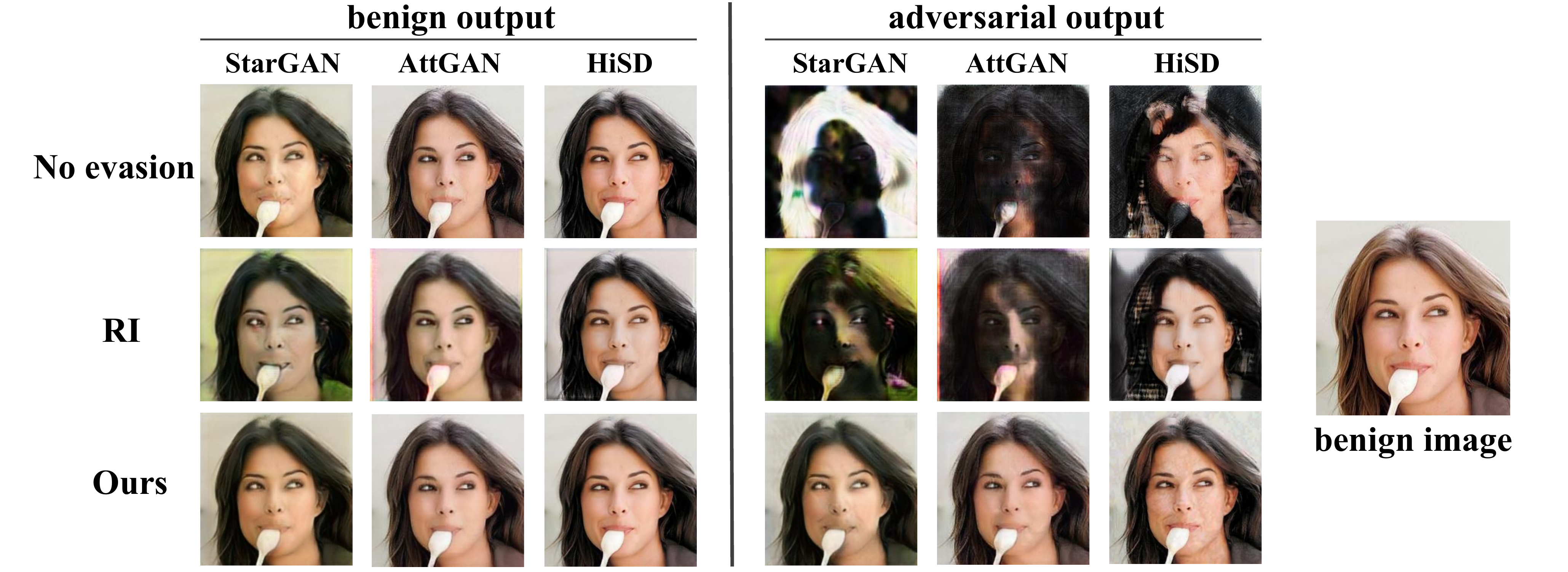}
      \caption{Visualization results in the leakage scenario. Traditional preprocessing-based adversarial defense methods fail in this scenario, but LoRA patching still shows good performance.}
      \label{fig:leakage_vis}
\end{figure}

In the leakage scenario, defenders are assumed to have prior access to the bypassing method and its parameters. Both quantitative (Fig.~\ref{fig:leakage}) and visual results (Fig.~\ref{fig:leakage_vis}) indicate that most baselines fail, as their preprocessing modules are integrated into the target model during adversarial perturbation generation. In contrast, LoRA Patching consistently maintains strong performance by leveraging an adversarial training–based paradigm. This design provides substantial robustness and demonstrates the stability and reliability of our method under challenging adversarial conditions.

\subsubsection{Comparison with Adversarial Training}
Table~\ref{tab:with_AT} presents a quantitative comparison of adversarial training-based methods, including our LoRA patching, in bypassing Disrupting~\cite{ruiz2020disrupting}.
Although MaGAT~\cite{luo2024magat} also achieves good performance, it suffers from substantial computational overhead. In contrast, our method requires only 4.6\% of its training time and 4.3\% of its parameter scale, since LoRA patching fine-tunes only a set of LoRA blocks rather than retraining the full Deepfake model. This efficiency is achieved by fully leveraging the expressive generative capacity of the pre-trained backbone.

\vspace{-12pt}

\subsection{Defensive LoRA Patching}

\begin{table}[t]
    \centering
    \caption{Quantitative results of defensive LoRA patching.}
    \label{tab:defensive_lora}
    \resizebox{1.0\columnwidth}{!}{
    \begin{tabular}{c|cccccc}
    \toprule
    \multirow{2}{*}{output} 
    &\multicolumn{2}{c}{StarGAN\cite{choi2018stargan}}  &\multicolumn{2}{c}{AttGAN\cite{he2019attgan}} 
    &\multicolumn{2}{c} {HiSD\cite{li2021image}} \\
      \cmidrule(l){2-3}  \cmidrule(l){4-5}  \cmidrule(l){6-7} & 
      FID$\downarrow$ &BRISQUE$\downarrow$ 
      &FID$\downarrow$ &BRISQUE$\downarrow$  
      &FID$\downarrow$ &BRISQUE$\downarrow$ \\
    \midrule
    Benign output &14.595 &17.483 &8.302 &20.280 &9.900 &17.224\\
    Adversarial output &14.860 &17.754 &7.941 &19.181 &9.588 &15.392\\      
    \bottomrule
    \end{tabular}}
\end{table}

\begin{figure}[t]
      \centering
      \includegraphics[width=1.0\linewidth]{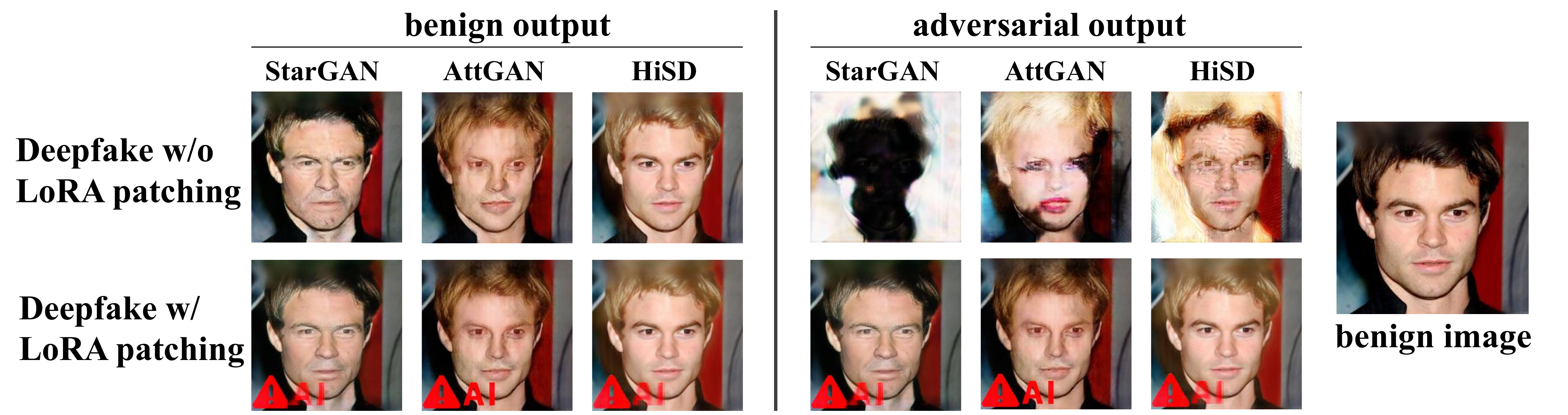}
      \caption{Visualization results of defensive LoRA patching. It embeds visible warnings in deepfake output.}
      \label{fig:defensive_lora}
\end{figure}

Table~\ref{tab:defensive_lora} reports the quantitative evaluation, with all FID scores below 15, indicating high visual quality. Fig.~\ref{fig:defensive_lora} shows that the visible warning ``AI'' is successfully embedded in both benign and adversarial outputs, demonstrating that defensive LoRA patching serves as a security-oriented complement to the base method. 
%We recommend that the Deepfake model publisher apply this patch.

\subsection{Ablation Study}

\begin{figure}[t]
      \centering
      \setcounter{subfigure}{0}
      \subfigure [Training loss of StarGAN.] {\includegraphics[width=.23\textwidth]{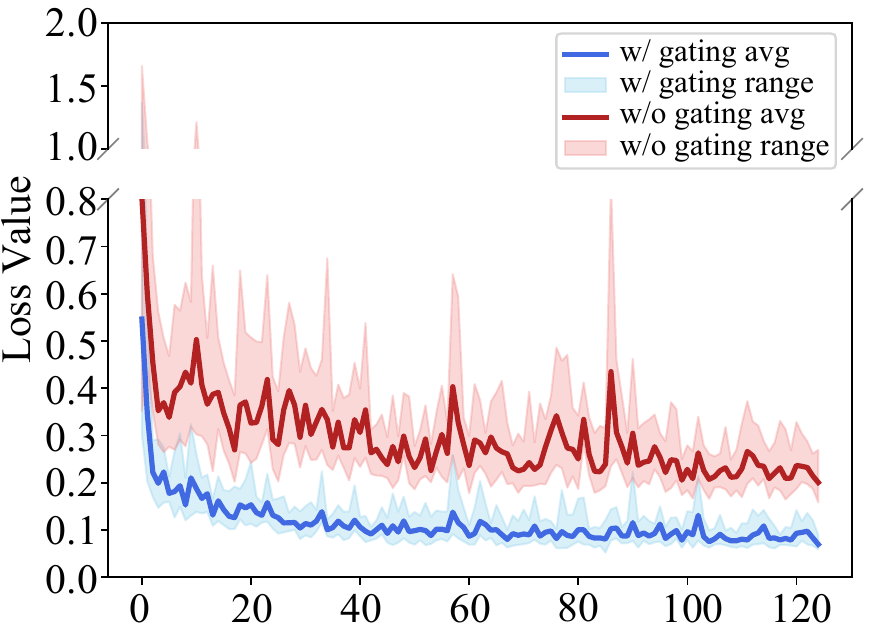}}
    \subfigure [Training loss of AttGAN.] {\includegraphics[width=.23\textwidth]{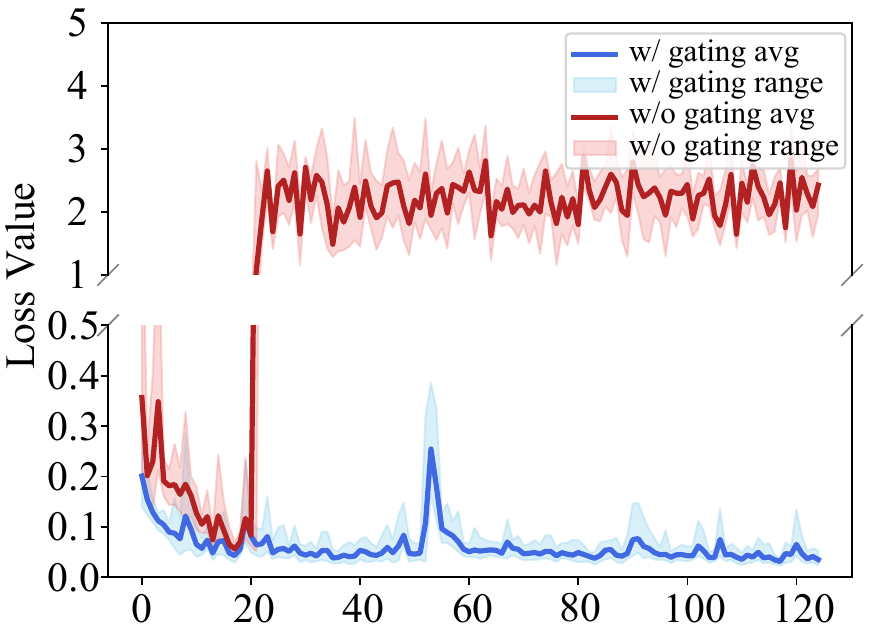}}
      \caption{The effect of using a gating mechanism on the fine-tuning process. The figure shows how the loss function changes with the number of iterations (every 10 iterations).}
      \label{fig:gating}
\end{figure}

\begin{table}[t]
    \centering
    \caption{Quantitative evaluation of MMFA Loss in defensive LoRA patching.}
    \label{tab:MMFA}
    \resizebox{1.0\columnwidth}{!}{
    \begin{tabular}{c|cccccc}
    \toprule
    \multirow{2}{*}{MMFA} 
    &\multicolumn{2}{c}{StarGAN\cite{choi2018stargan}}  &\multicolumn{2}{c}{AttGAN\cite{he2019attgan}} 
    &\multicolumn{2}{c} {HiSD\cite{li2021image}} \\
      \cmidrule(l){2-3}  \cmidrule(l){4-5}  \cmidrule(l){6-7} & 
      SSIM$\uparrow$ &CLIP score$\uparrow$ 
      &SSIM$\uparrow$ &CLIP score$\uparrow$  
      &SSIM$\uparrow$ &CLIP score$\uparrow$ \\
    \midrule
    w/o          
    &0.739 &0.896 &0.923 &0.857 &0.835 &0.779\\
    w/         
    &0.868 &0.931 &0.926 &0.963 &0.894 &0.929\\      
    \bottomrule
    \end{tabular}}
\end{table}

\begin{table}[t]
    \centering
    \caption{Quantitative results of different rank settings.}
    \label{tab:rank_of_lora}
    \resizebox{1.0\columnwidth}{!}{
    \begin{tabular}{c|ccccccccc}
    \toprule
    \multirow{2}{*}{Rank} 
    &\multicolumn{3}{c}{StarGAN\cite{choi2018stargan}}  &\multicolumn{3}{c}{AttGAN\cite{he2019attgan}} 
    &\multicolumn{3}{c} {HiSD\cite{li2021image}} \\
      \cmidrule(l){2-4}  \cmidrule(l){5-7}  \cmidrule(l){8-10}& 
      $L_{2}$$\downarrow$ &SSIM$\uparrow$  &DSR$\downarrow$ 
    &$L_{2}$$\downarrow$ &SSIM$\uparrow$  &DSR$\downarrow$ 
    &$L_{2}$$\downarrow$ &SSIM$\uparrow$  &DSR$\downarrow$ \\
    \midrule
    4          
    &0.034 &0.867 &0.162 &0.011 &0.919 &0.014 &0.021 &0.899 &0.120\\
    8          
    &0.017 &0.917 &0.026 &0.007 &0.935 &0.006 &0.010 &0.945 &0.017\\
    16     
    &0.015 &0.921 &0.016 &0.008 &0.938 &0.006 &0.009 &0.947 &0.010\\
    32     
    &0.014 &0.923 &0.011 &0.006 &0.939 &0.006 &0.008 &0.947 &0.008\\
    \bottomrule
    \end{tabular}}
\end{table}

\subsubsection{Gating Mechanism}
As shown in Fig.~\ref{fig:gating}, the gating mechanism promotes convergence of StarGAN’s LoRA patch and prevents gradient explosion during AttGAN fine-tuning. It does so by adaptively regulating each LoRA block’s contribution, avoiding overly large updates that could destabilize training.

\subsubsection{MMFA}
We focus on the effect of the MMFA Loss for defensive LoRA patching, as the visible warning carries rich semantic information. CLIP scores~\cite{radford2021learning} are used to assess semantic alignment between generated images and $y_w$. As shown in Table~\ref{tab:MMFA}, MMFA Loss improves semantic consistency and overall image quality, increasing CLIP scores by over 10\% due to the inclusion of semantic features.

\subsubsection{Rank of LoRA}
We evaluate the effect of rank on bypassing proactive defense. As shown in Table~\ref{tab:rank_of_lora}, increasing the rank improves performance, as higher-rank LoRA matrices can capture more complex adversarial perturbation patterns. However, larger ranks incur higher computational cost, allowing a trade-off between performance and efficiency.

\section{Conclusion}

We reveal a significant limitation in current proactive Deepfake defenses and propose LoRA patching as an effective defense bypass. By integrating a plug-and-play LoRA patch with a learnable gating mechanism and MMFA loss, our method preserves high-quality outputs while resisting adversarial watermarks. Extensive experiments show strong performance with minimal overhead. Defensive LoRA patching further adds visible warnings as a complementary measure. We hope this work will encourage the community to develop more robust defenses against Deepfakes, mitigating this security concern.

\clearpage
\bibliographystyle{IEEEtran}
\balance
\bibliography{ref}

\end{document}